# Image Matching Using SIFT, SURF, BRIEF and ORB: Performance Comparison for Distorted Images

Ebrahim Karami, Siva Prasad, and Mohamed Shehata

Faculty of Engineering and Applied Sciences, Memorial University, Canada

*Abstract*-Fast and robust image matching is a very important task with various applications in computer vision and robotics. In this paper, we compare the performance of three different image matching techniques, i.e., SIFT, SURF, and ORB, against different kinds of transformations and deformations such as scaling, rotation, noise, fish eye distortion, and shearing. For this purpose, we manually apply different types of transformations on original images and compute the matching evaluation parameters such as the number of key points in images, the matching rate, and the execution time required for each algorithm and we will show that which algorithm is the best more robust against each kind of distortion.

## I. INTRODUCTION

Feature detection is the process of computing the abstraction of the image information and making a local decision at every image point to see if there is an image feature of the given type existing in that point. Feature detection and image matching have been two important problems in machine vision and robotics, and their applications continue to grow in various fields. An ideal feature detection technique should be robust to image transformations such as rotation, scale, illumination, noise and affine transformations. In addition, ideal features must be highly distinctive, such that a single feature to be correctly matched with high probability [1, 2].

Scale Invariant Feature Transform (SIFT) is a feature detector developed by Lowe in 2004 [3]. Although SIFT has proven to be very efficient in object recognition applications, it requires a large computational complexity which is a major drawback especially for real-time applications [3, 4]. There are several variants and extension of SIFT which have improved its computational complexity [5-7].

Speed up Robust Feature (SURF) technique, which is an approximation of SIFT, performs faster than SIFT without reducing the quality of the detected points [8]. Both SIFT and SURF are thus based on a descriptor and a detector. Binary Robust Independent Elementary Features (BRIEF) is another alternative for SIFT which requires less complexity than SIFT with almost similar matching performance [9]. Rublee et al. proposed Oriented FAST and Rotated BRIEF (ORB) as another efficient alternative for SIFT and SURF [10]. Any of these feature detection methods can also be employed in remote sensing application such as sea ice applications, e.g. Zehen Lieu et al. used the SIFT algorithm based matching to find the icebergs whose shapes has changed due to collision or splits [11]. Also feature tracking algorithms are used for ice motion tracking e.g. Ronald Kwok [12]. There are a few works available on the comparison of SIFT and SURF [13-15] and from the best of our knowledge, there are no papers on the comparison of ORB with them. In this paper, we compare the performance of SIFT, SURF, and ORB techniques. We also compare the robustness of these techniques against rotation, scaling, and deformity due to horizontal or vertical shears, and fish eye. Fish eye distortions are used for creating hemispherical panoramic images. There can be caused by lens of camera or manually created by using spherical distortions. Planetariums use the fish eye paperion of night sky, flight simulations in order to create immersive environment for the trainee's uses the fish eye paperion, some motion-picture formats also uses these paperions [16]. In meteorology fish eye lens are used to capture cloud formations. We also aim to find required number of points of interest in each case. Fig. 1 show an example of image which was subject to fisheye distortion.

The rest of this paper is organized as follows: Section II presents an overview for the three image matching techniques SIFT, SURF, and ORB. In Section III, we investigate the sensitivity of SIFT, SURF, and ORB against each intensity, rotation, scaling, shearing, fish eye distortion, and noise. Section IV concludes the paper.

## II. OVERVIEW OF IMAGE MATCHING TECHNIQUES

### SIFT

SIFT proposed by Lowe solves the image rotation, affine transformations, intensity, and viewpoint change in matching features. The SIFT algorithm has 4 basic steps. First is to estimate a scale space extrema using the Difference of Gaussian (DoG). Secondly, a key point localization where the key point candidates are localized and refined by eliminating the low contrast points. Thirdly, a key point orientation assignment based on local image gradient and lastly a descriptor generator to compute the local image descriptor for each key point based on image gradient magnitude and orientation [3].

*SURF*

SURF approximates the DoG with box filters. Instead of Gaussian averaging the image, squares are used for approximation since the convolution with square is much faster if the integral image is used. Also this can be done in parallel for different scales. The SURF uses a BLOB detector which is based on the Hessian matrix to find the points of interest. For orientation assignment, it uses wavelet responses in both horizontal and vertical directions by applying adequate Gaussian weights. For feature description also SURF uses the wavelet responses. A neighborhood around the key point is selected and divided into subregions and then for each subregion the wavelet responses are taken and represented to get SURF feature descriptor. The sign of Laplacian which is already computed in the detection is used for underlying interest points. The sign of the Laplacian distinguishes bright blobs on dark backgrounds from the reverse case. In case of matching the features are compared only if they have same type of contrast (based on sign) which allows faster matching [7].

*ORB*

ORB is a fusion of the FAST key point detector and BRIEF descriptor with some modifications [9]. Initially to determine the key points, it uses FAST. Then a Harris corner measure is applied to find top N points. FAST does not compute the orientation and is rotation variant. It computes the intensity weighted centroid of the patch with located corner at center. The direction of the vector from this corner point to centroid gives the orientation. Moments are computed to improve the rotation invariance. The descriptor BRIEF poorly performs if there is an in-plane rotation. In ORB, a rotation matrix is computed using the orientation of patch and then the BRIEF descriptors are steered according to the orientation.

## III. SIMULATION RESULTS

In this Section, we investigate the sensitivity of SIFT, SURF, and ORB against each intensity, rotation, scaling, shearing, fish eye distortion, and noise.

*Intensity*

The images with varying intensity and color composition values were used to compare the algorithms and results are presented in Table 1 and Figure 1.

For images with varying intensity values, SIFT provides the best matching rate while ORB has the least. Computational time requirement for ORB is the least.

*Rotation*

We considered here a rotation of 45 degree to the image to be matched. The results are given in the Table 2 and Figure 2. With rotated image, as one can see from Table 2, SIFT provides a 65 % matching rate. Table 3 presents the matching rate for different rotation angles. From it, one can see that with rotation angles proportional to 90 degree, ORB and SURF always present the best matching rate, while for other angles of rotations such as 45, 135, and 225, SIFT presents the highest matching rate.

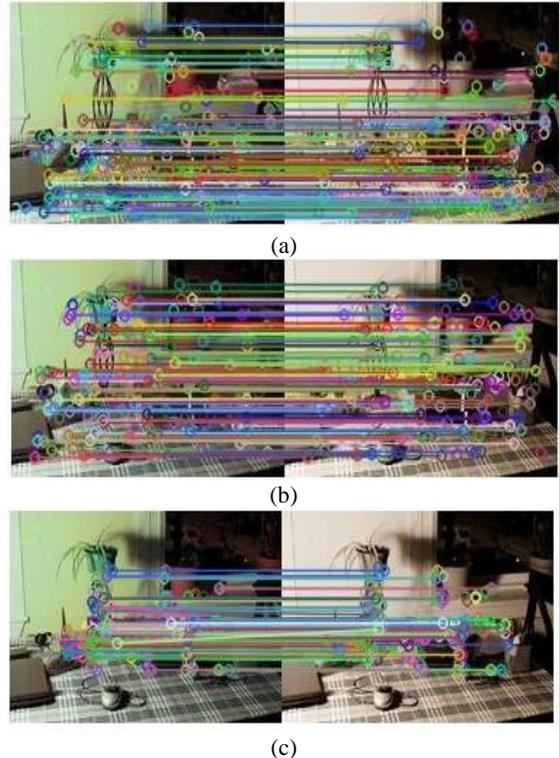

Figure 1. The matching of varying intensity images using (a) SIFT (b) SURF (c) ORB.

Table 1. Results of comparing the images with varying intensity.

|      | Time (sec) | Kpnts1 | Kpnts2 | Matches | Match rate (%) |
|------|------------|--------|--------|---------|----------------|
| SIFT | 0.13       | 248    | 229    | 183     | 76.7           |
| SURF | 0.04       | 162    | 166    | 119     | 72.6           |
| ORB  | 0.03       | 261    | 267    | 168     | 63.6           |

Table 2. Results of comparing the image with its rotated image.

|      | Time (sec) | Kpnts1 | Kpnts2 | Matches | Match rate (%) |
|------|------------|--------|--------|---------|----------------|
| SIFT | 0.16       | 248    | 260    | 166     | 65.4           |
| SURF | 0.03       | 162    | 271    | 110     | 50.8           |
| ORB  | 0.03       | 261    | 423    | 158     | 46.2           |

Table 3. Matching rate versus the rotation angle.

| Angle → | 0   | 45 | 90 | 135 | 180 | 225 | 270 |
|---------|-----|----|----|-----|-----|-----|-----|
| SIFT    | 100 | 65 | 93 | 67  | 92  | 65  | 93  |
| SURF    | 99  | 51 | 99 | 52  | 96  | 51  | 95  |
| ORB     | 100 | 46 | 97 | 46  | 100 | 46  | 97  |

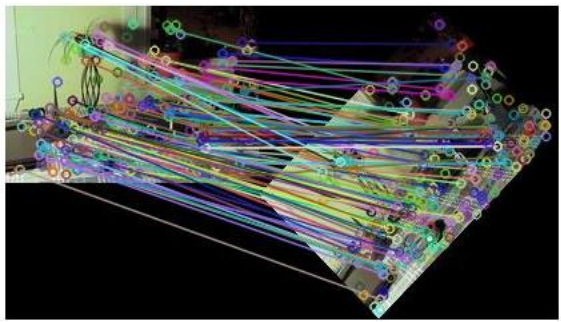

(a)

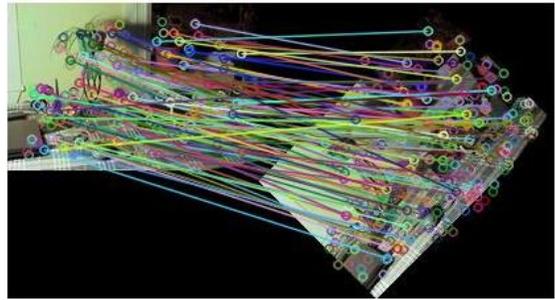

(b)

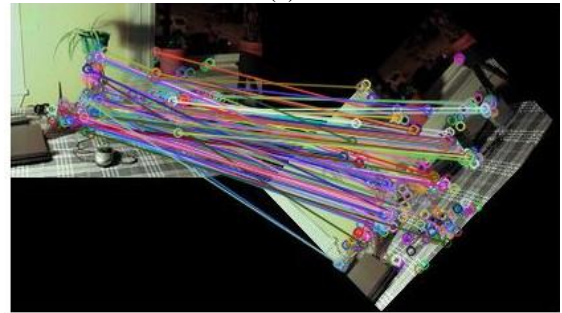

(c)

Figure 2. The matching of the original image with its rotated image using: (a) SIFT (b) SURF (c) ORB.

Table 4. Results of comparing the image with its scaled image.

|  | Time (sec) | Kpnts1 | Kpnts2 | Matches | Match rate (%) |
|---|---|---|---|---|---|
| SIFT | 0.25 | 248 | 1210 | 232 | 31.8 |
| SURF | 0.08 | 162 | 581 | 136 | 36.6 |
| ORB | 0.02 | 261 | 471 | 181 | 49.5 |

Table 5. Results of comparing the image with its sheared image.

|  | Time (sec) | Kpnts 1 | Kpnts 2 | Matches | Match rate (%) |
|---|---|---|---|---|---|
| SIFT | 0.133 | 248 | 229 | 150 | 62.89 |
| SURF | 0.049 | 162 | 214 | 111 | 59.04 |
| ORB | 0.026 | 261 | 298 | 145 | 51.88 |

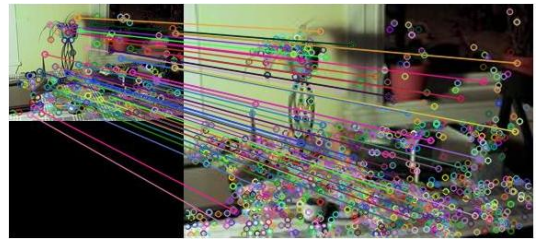

(a)

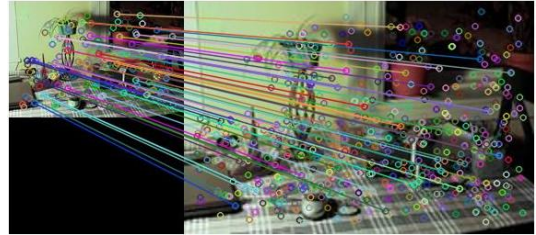

(b)

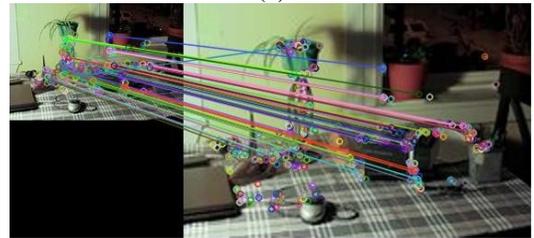

(c)

Figure 3. The matching of the original image with its scaled image using: (a) SIFT (b) SURF (c) ORB.

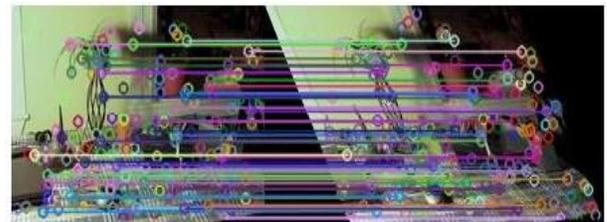

(a)

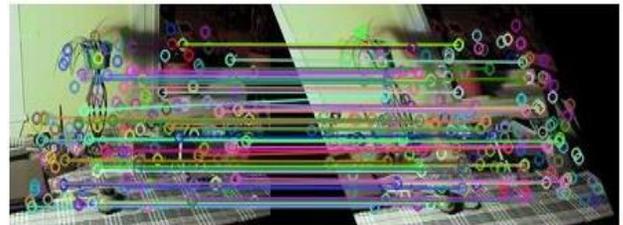

(b)

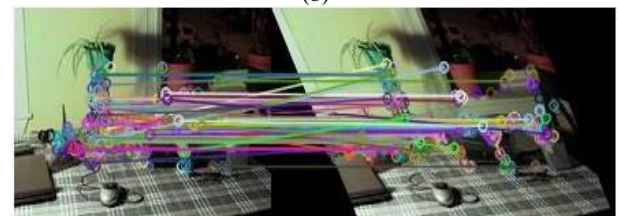

(c)

Figure 4. The matching of an image with its sheared image using: (a) SIFT (b) SURF (c) ORB.

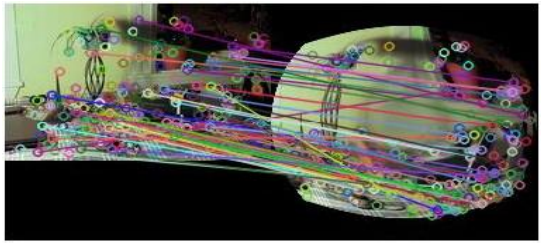
(a)

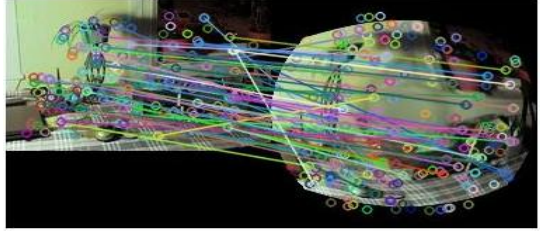
(b)

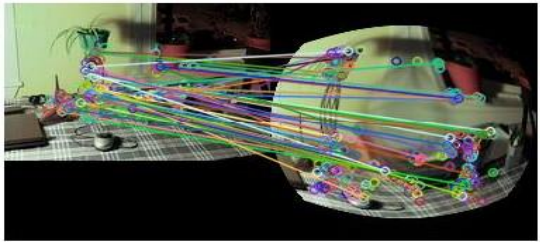
(c)

Figure 5. The matching of an image with its fisheye distorted image using: (a) SIFT (b) SURF (c) ORB.

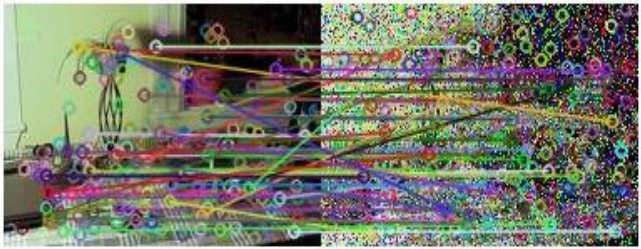
(a)

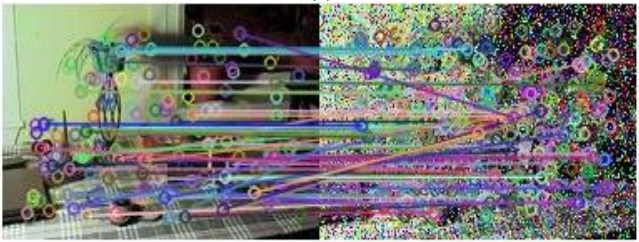
(b)

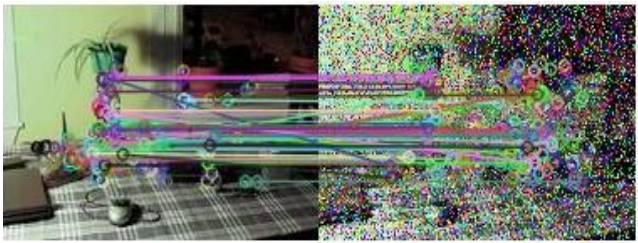
(c)

Figure 6. The matching of image with the image added with a salt and pepper noise using (a) SIFT (b) SURF (c) ORB.

Table 6. Results of comparing the image with its fish eye distorted image.

|  | Time (sec) | Kpnts 1 | Kpnts 2 | Matches | Match rate (%) |
|---|---|---|---|---|---|
| SIFT | 0.132 | 248 | 236 | 143 | 59.09 |
| SURF | 0.036 | 162 | 224 | 85 | 44.04 |
| ORB | 0.012 | 261 | 282 | 125 | 46.04 |

Table 7. Results of the image matching by adding 30 % of salt and pepper noise.

|  | Time (sec) | Kpnts1 | Kpnts2 | Matches | Match rate (%) |
|---|---|---|---|---|---|
| SIFT | 0.115 | 248 | 242 | 132 | 53.8 |
| SURF | 0.059 | 162 | 385 | 108 | 39.48 |
| ORB | 0.027 | 261 | 308 | 155 | 54.48 |

*Scaling*

In this scenario, the image was scaled by 2 times to see the effect of matching with respect to scaling. Results are shown in Table 4 and Figure 3. The highest matching rate is for ORB while the least is noticed for SIFT.

*Shearing*

In this scenario, the original image was sheared with value 0.5 to see the effect of matching with respect to shearing. The results are presented in Table 5 and Figure 4. From Table 5, one can see that the highest matching rate is achieved from SIFT.

*Fisheye distortion*

The results are presented in Table 6 and Figure 5. From Table 6, one can see that the highest matching rate is obtained from SIFT, and from Figure 5, one can see that there are relatively less correct matches compared to the previous scenarios.

*Noisy images*

In this case, 30 % salt and pepper noise is added to the original image to see the effect of noise on the matching rate. From Table 7 and Figure 6, one can see that ORB and SIFT shows the best matching rates. The results might slightly vary but SIFT and ORB provide the highest matching rates. The added salt and pepper noise is randomly distributed and hence may be affecting some of the key points, but both SIFT and ORB show almost equal performance.

## IV. CONCLUSION

In this paper, we compared three different image matching techniques for different kinds of transformations and deformations such as scaling, rotation, noise, fisheye distortion, and shearing. For this purpose, we applied different types of transformations on original images and displayed the matching evaluation parameters such as the number of key points in images, the matching rate, and the execution time required for each algorithm.

We showed that ORB is the fastest algorithm while SIFT performs the best in the most scenarios. For special case when the angle of rotation is proportional to 90 degrees, ORB and SURF outperforms SIFT and in the noisy images, ORB and SIFT show almost similar performances. In ORB, the features are mostly concentrated in objects at the center of the image while in SURF, SIFT and FAST key point detectors are distributed over the image.